%% file: iclr2021_conference.tex
\documentclass{article} 
\usepackage{iclr2021_conference,times}

\input{math_commands.tex}

\usepackage{hyperref}
\usepackage{url}
\usepackage{graphicx}
\usepackage{dsfont}

\title{Using system context information to complement weakly labeled data}

\author{Matthias Meyer \& Lothar Thiele \\
Computer Engineering and Networks Laboratory \\
ETH Zurich \\
Zurich, Switzerland\\
\texttt{\{matthmey,thiele\}@ethz.ch} \\
\And
Michaela Wenner \& Fabian Walter \\ 
Laboratory of Hydraulics, Hydrology and Glaciology \\
ETH Zurich/WSL Birmensdorf \\
\texttt{\{wenner,walter\}@vaw.baug.ethz.ch} \\
\AND
Clément Hibert \\
Institut de Physique du Globe de Strasbourg \\
University of Strasbourg\\
\texttt{hibert@unistra.fr}
}

\iclrfinalcopy 
\begin{document}

\maketitle

\begin{abstract}
Real-world datasets collected with sensor networks often contain incomplete and uncertain labels as well as artefacts arising from the system environment. Complete and reliable labeling is often infeasible for large-scale and long-term sensor network deployments due to the labor and time overhead, limited availability of experts and missing ground truth. In addition, if the machine learning method used for analysis is sensitive to certain features of a deployment, labeling and learning needs to be repeated for every new deployment. To address these challenges, we propose to make use of system context information formalized in an information graph and embed it in the learning process via contrastive learning. Based on real-world data we show that this approach leads to an increased accuracy in case of weakly labeled data and leads to an increased robustness and transferability of the classifier to new sensor locations.
\end{abstract}

\section{Introduction}

Classifiers based on artificial neural networks have proven to be very effective across domains, however their applicability to real-world data is limited by the requirement of a clean and comprehensive dataset \citep{tsiprasImageNetImageClassification2020}. 
Unfortunately, real-world datasets often contain artefacts arising from the system environment and contain incomplete and uncertain labels. One example of machine learning applications is natural hazard monitoring for slope failure detection \citep{Hammer2013, Dammeier2016}. Here, high misclassification requires careful retraining and post-processing \citep{Hibert2017}. In this setting, comprehensive manual annotations are infeasible for large-scale and long-term sensor network deployments due to the labor and time overhead \citep{meyerSystematicIdentificationExternal2019b}. 

Hence, the process is error-prone and requires significant domain expertise. However, experts might not be available throughout the whole deployment periods of the sensor network, which inevitably leads to an annotation set containing noisy annotations limited in time and/or subset of sensors. In addition, as long as the learned features and classifier are sensitive to the detailed properties of the subsurface and the sensors, 
labeling and learning needs to be repeated for every new installation or classifier performance is decreased \citep{Wenner2020}. Therefore, there is a close link between weakly labeled data and robustness with respect to certain feature variations.

Fortunately, real-world deployments provide additional sources of information which could be beneficial for learning, such as correlation of sensor data due to sensor proximity. However, these information cannot be easily captured by the prevailing data/annotation pairs used for learning. Similarity learning \citep{schroffFaceNetUnifiedEmbedding2015a,meyerUnsupervisedFeatureLearning2017}, such as contrastive learning \citep{heMomentumContrastUnsupervised2020,chenSimpleFrameworkContrastive2020,saeedContrastiveLearningGeneralPurpose2020} allows to establish relations between data pairs. However, their capability to integrate system context information is limited.

To address these challenges, we propose to transfer the concept of knowledge graphs \citep{hoganKnowledgeGraphs2021} to learning by using it for storing information about data similarity. Moreover, we extend the prevailing data/annotation learning concept to allow any data point to be an annotation for any other data point. This is accomplished by utilizing the following concepts: (i) injecting all available knowledge in form of an information graph and sampling from it, (ii) transforming the data into a common representation and (iii) the use of contrastive learning to train the system. We show that using these concepts to formalize system context information and using the additional knowledge in the learning phase leads to an increased accuracy in case of weakly labeled data and leads to an increased robustness and transferability of the classifier to new sensor locations. \footnote{Further content is available at https://matthiasmeyer.xyz/system-context-info/}


Our main contributions are:
\begin{itemize}
    \item We present a method which uses system context information to counteract the negative impact of few and weak labels by combining contrastive learning with an information graph.
    \item We present a unified learning process in which annotations are encoded as Gaussian random vectors to treat them similar to data.
    \item We demonstrate on a dataset gathered from a real-world deployment in the Swiss alps, how the method can be used to train a classifier with improved generalization performance across sensors with diverging characteristics.
\end{itemize}

\section{Dataset}

In this work, we use data from a real-world deployment of seismic sensors at Illgraben, Switzerland. The sensor array consists of 8 seismometer (ILL01-08), each having three channels, one vertical and two horizontal. The sensors are deployed at distances of hundreds of meters up to several kilometers away from the area of interest. We aim to distinguish seismic signals from 3 different types of events namely earthquakes, slope failures and noise signals. The Illgraben event catalog was created by visual inspection of the vertical channel of the seismic waveforms and their spectrograms by experts. The earthquake catalogs provided by the Swiss Seismological Service (SED) and the European-Mediterranean Seismological Center (EMSC) served as additional ground truth for providing correct earthquake labels. 
The Illgraben event catalog consists of 320 to 560 time segments per station each containing an event, summing up to 32.5 hours of labelled seismic data recorded at a sampling frequency of 100 Hz. 
In addition, the dataset contains randomly sampled, verified time segments without activity with a total duration approximately equal to the event segment's total duration. 

\section{Method}
In our scenario, two major issues need to be addressed, namely (i) few and weak labels and (ii) classifier robustness. The first issue requires an improvement and extension of the annotation set, the latter requires that the learning method can adapt to out-of-distribution samples.

In contrast to real-time classification, environmental monitoring usually relies on post-processing of a long-term dataset. Therefore, an extensive dataset is usually available for training albeit not always thoroughly annotated. 
In our scenario, we can make use of non-annotated data by using general assumptions about the specific sensor deployment, for example about sensor proximity: The same event is captured by multiple seismometer channels and possibly multiple stations, but with different signal signatures. These differences are caused by groundwave propagation as well as properties of the seismometers, for example ground coupling. Thus, we obtain "different views" of the same event. Contrastive learning has shown to benefit from such different views.
Intuitively speaking, contrastive learning achieves an embedding of data samples in a latent space by moving representations of different views of the same event closer together while increasing the distances of representations of different events. 

To combine contrastive learning with all available information, we propose to make use of an information graph, which holds annotations as well information about the relation between time segments, channels and stations.
We expect that by using system context information we simultaneously (i) improve our annotation set by adding additional information to each annotated segment and (ii) include non-annotated, out-of-distribution samples in the training process.

The information graph is filled by subdividing the seismic signals into segments using a window length $T_w$. Each segment is represented by a node in the information graph. An edge is introduced between segments A and B if the segments overlap in time and A is from a different station or different channel than segment B. To reduce the possibility of learned shortcuts \citep{fonsecaUnsupervisedContrastiveLearning2020} no edges to segments of the same channel are added.

\begin{figure}[h]
\begin{center}
    \includegraphics[width=0.8\linewidth]{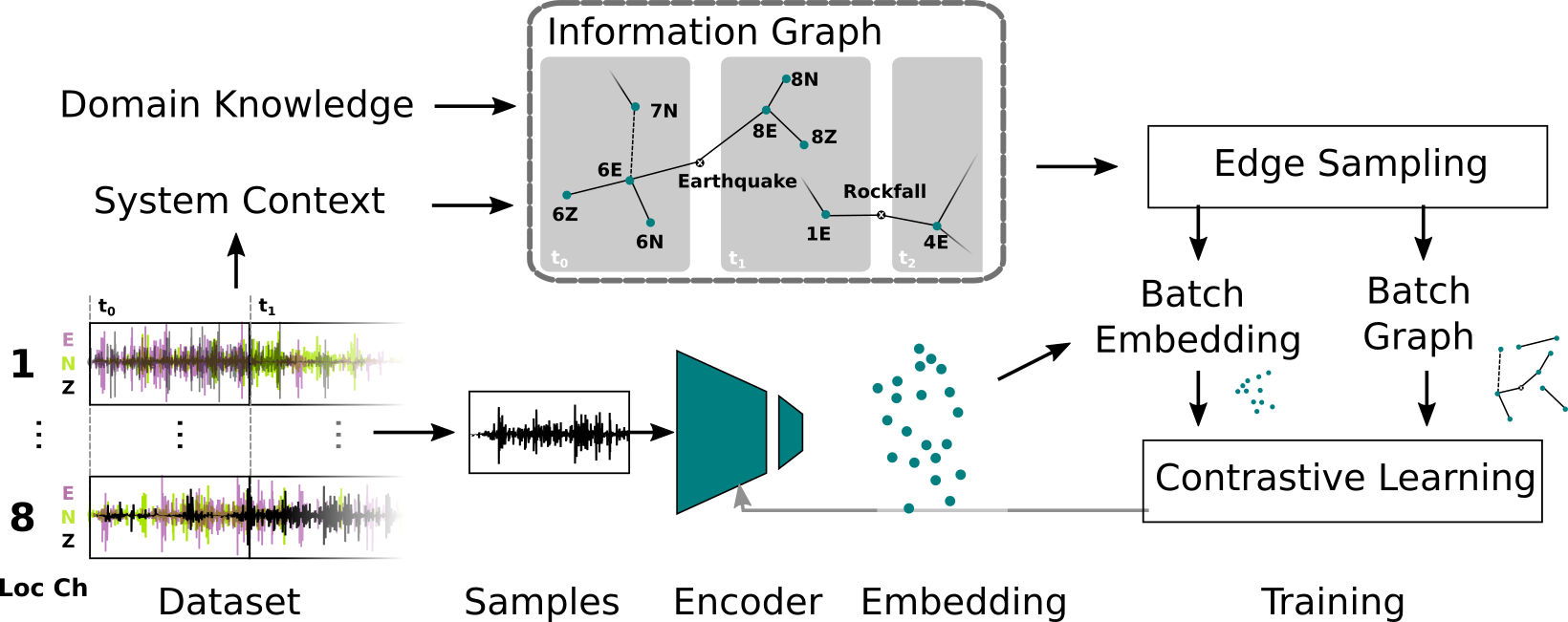} 
\end{center}
\caption{The central entity is the information graph which combines knowledge from domain experts, for example annotations or signal propagation behaviour, as well as dataset-specific knowledge of each data segment, for example \textbf{loc}ation, \textbf{ch}annel, time. The combination of contrastive loss, information graph and encoder allows to learn a suitable embedding for the classification task. }\label{fig:procedure}
\end{figure}

The information graph is used to train a model $f(\cdot)$ which embeds each segment $x_i$ into a common space $z_i=f(x_i)$, with $\vz_i \in \R^d$. Similar to related work, we separate the model into an encoder and encoder head \citep{chenSimpleFrameworkContrastive2020}. As illustrated in Fig.\,\ref{fig:procedure}, a fixed number $N_e$ of edges is sampled from the graph for every batch during training and connected data segments are loaded. Any duplicate segments are removed from the batch before computing $f(\cdot)$, leading to a number of data segments in the batch of $N \leq 2N_e$. Each data segment is encoded by the encoder and subsequently transformed by the encoder head into an embedding vector $\vz_i$.

By sampling the edges we construct a subgraph of the information graph with non-negative adjacency matrix $\displaystyle \mA \in \R_{\geq0}^{N \times N}$. To avoid that second order neighbours of a node have a detrimental impact on training by not being directly connected, 
we add second order neighbors by $\mB = \mA + \mA^2$. In this setting, we define the contrastive loss between a pair $s,t$ (source and target of an edge), with the adjacency matrix $\mB$ as follows:

\begin{equation} \label{eq:1}
\displaystyle L_{s,t} = -\emB_{s,t}\log \frac{\exp(\phi(\vz_s,\vz_t)/\tau)}{\sum_{n=1}^{N} \mathds{1}_{[B_{s,n}=0]} \exp(\phi(\vz_s,\vz_n)/\tau) }
\end{equation}

where $\emB_{s,t}$ represents the weight of the edge connecting $s,t$. $\phi(\cdot)$ is a similarity function, which in our implementation is the Cosine similarity. $\tau$ is a temperature scaling. The indicator function $\mathds{1}$ is evaluating to 1 iff $\displaystyle \emB_{s,n}$ is zero.

Annotations are considered in the information graph by introducing $N_c$ anchor nodes, where $N_c$ equals the number of classes. Each segment belonging to a class is connected to the anchor node of that class by an edge. Two strategies can be employed to compute Eq. \ref{eq:1} for an edge with an annotation anchor node. 

The first option makes use of the fact that $\mB$ contains second order neighbors meaning that all nodes sharing an edge with an annotation are also directly connected. Thus, the edges with an annotation node can be skipped while computing Eq. \ref{eq:1} but representations with the same annotation are still moved closer together, which resembles the work by \citet{khoslaSupervisedContrastiveLearning2020a}. 

The second option is to introduce a high-dimensional L2-normalized Gaussian random vector $\va^{(c)} \in \R^d$ for class $c$ into the batch which acts as the target $z_t$ during computation of Eq. \ref{eq:1}. The vectors are fixed at the beginning of the training. Here, we make use of the fact that any two random high-dimensional vectors are almost orthogonal to each other with high probability \citep{blumFoundationsDataScience2020}, thus data points of different classes are trained to move "far away" from each other. The first strategy will be referred to as \textit{link}, the latter as \textit{anchor} in the following evaluation.

\section{Experimental evaluation}

We evaluate the proposed approach by performing an ablation study and comparing the system to a random forest classifier, which is best practice for slope failure detection \citep{Wenner2020}. 

For the ablation study we use a classifier based on a single-channel variant of ResNet18 \citep{heDeepResidualLearning2015} as encoder in combination with an MLP with 1 hidden layer as encoder head. During classification the encoder head is replaced with a classification head differing only in output size. Input to the ResNet18 is a log-compressed spectrogram of the seismic data. For more implementation details please refer to the Appendix \ref{implementation_details}. The ResNet18 model is trained with three methods, cross-entropy loss between the output of the classification head and the ground truth (Resnet18+XE), contrastive pretraining using the information graph (IG) and either the link (Resnet18+IG+link) or anchor (Resnet18+IG+anchor) strategy. Subsequently, the classification head is trained using cross-entropy loss. We compare training with system context information (SC) and training without it. 

The benefit of our approach for the weakly-labeled setting is evaluated by using the available training annotations of all stations. The experiments are repeated 5 times and mean and standard deviation are reported. Robustness is evaluated by training the model variants when only a subset of the annotations are available. While the whole training data is available to train a classifier only the annotations for one of each of the 8 seismic stations can be used. All reported accuracies are based on evaluation on the test set using all stations and are reported as mean and standard deviation of all one-station evaluations.

The results presented in Table \ref{tab:results} show that the ResNet18 classifiers outperform the random forest classifier in the weakly-labeled setting (\textit{all} and \textit{all+SC}). The \textit{all} column, illustrates that training using contrastive pretraining improves the performance significantly in comparison to the random forest classifier but only slightly in comparison to using cross-entropy loss (ResNet18+XE). However, if we include the system context information the accuracy improves significantly (\textit{all+SC} column), demonstrating our method's applicability to weakly labeled data.

\begin{figure}[h]
\begin{center}
    \includegraphics[width=0.9\linewidth]{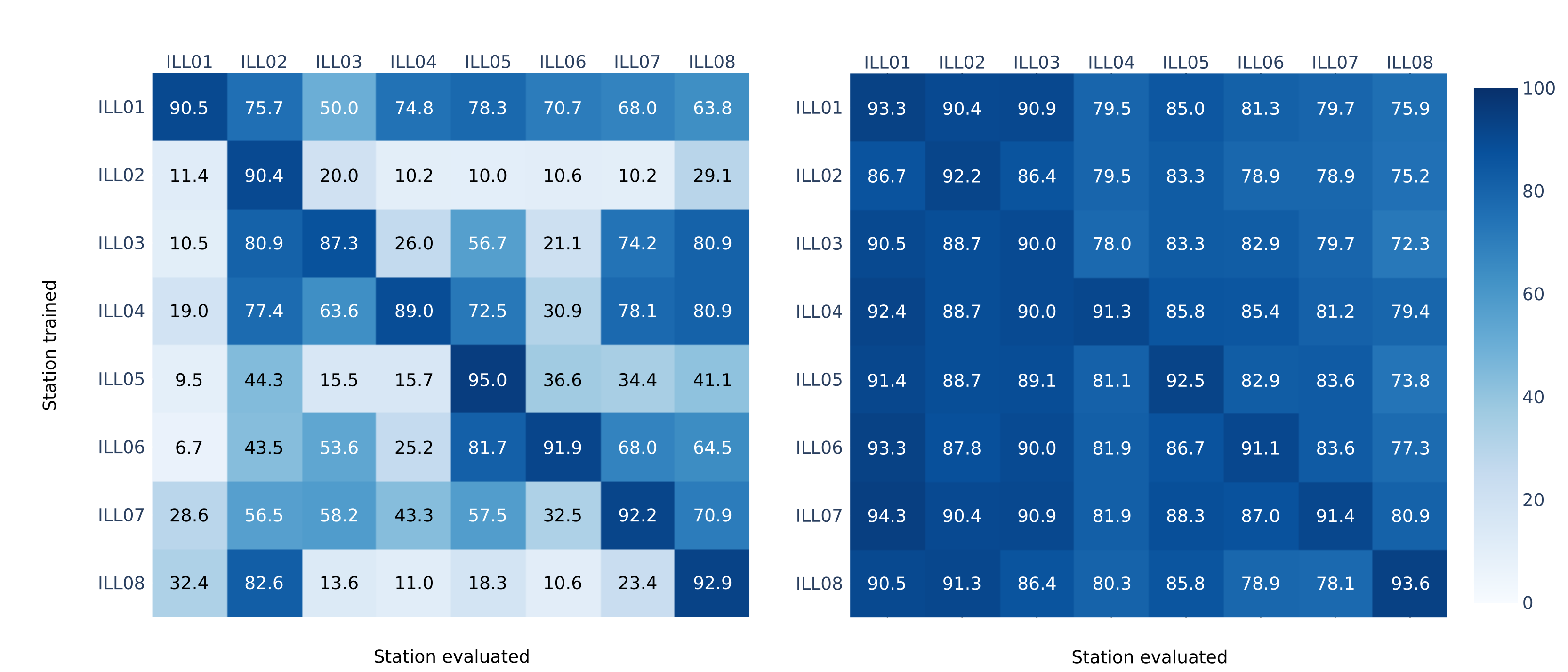}
\end{center}
\caption{Each row indicates of which station the annotated training subset was used. Each column indicates the respective score on the subset of the test dataset for each station. (Left): Results for ResNet18+XE. (Right): Results for ResNet18+IG+anchor.}\label{fig:acc_matrix}
\end{figure}

In the robustness experiments, all classifiers show a comparable bad performance of around 50\,\% on average if only annotated data of one station is used (\textit{one-station}). If more non-annotated data from other stations is available, our method takes advantage of the system context information (SC) stored in the information graph (\textit{one-station+SC}) and the average accuracy rises to over 84\,\%. The increase comes from a better generalization to other sensors, as illustrated in Fig.\,\ref{fig:acc_matrix}. The left figure illustrates the poor generalization of ResNet18+XE to other stations than the one used for training. If, however, non-annotated data from other stations and system context information is available, our method increases classifier performance on all stations, thus demonstrating and increased robustness.

\begin{table}[]
    \centering
    \begin{tabular}{l|lllll}
          & \multicolumn{3}{c}{Accuracy} \\ 
         & all & all+SC & one-station & one-station+SC \\ 
        \hline
        Random Forest & 86.4 \% & - & n.a. & -  \\
        ResNet18+XE & 91.3 \%  $\pm$ 0.5 & - & 50.0 \% $\pm$ 15.0 & - \\ 
        ResNet18+IG+links &   92.3 \% $\pm$ 0.3  & 93.8 $\pm$ 0.3  & 44.0 \% $\pm$ 15.6 & 84.1 \% $\pm$ 2.6  \\ 
        ResNet18+IG+anchors & 92.0 \% $\pm$ 0.4 & 93.9 $\pm$ 0.6 & 47.9 \% $\pm$ 16.7 & 85.0 \% $\pm$ 1.8 & \\
    \end{tabular}
    \caption{Classifier accuracies for different sets of available training annotations. Either all annotations (\textit{all}) or only annotations for one station are available (\textit{one-station}). Additionally, the information graph (IG) is used with or without system context information (\textit{SC}). }
    \label{tab:results}
\end{table}

\newpage
\section{Conclusion}
In this paper we have presented a novel approach to learn with weakly labeled data for the case of mass movement monitoring. By using contrastive learning we can increase the classification accuracy compared to the reference implementation. Moreover, the presented method unifies data and annotation representations and thus inherently allows to integrate additional system information into the learning process. This additional information leads to a strong performance increase in a setting with limited annotations and diverging sensor characteristics, demonstrating increased robustness across sensors.

\newpage
\bibliography{iclr2021_conference}
\bibliographystyle{iclr2021_conference}

\newpage
\appendix
\section{Implementation Details}
\label{implementation_details}
\subsection{Details to Contrastive Learning with Information Graph}

The seismic signals are subdivided into segments using a window length $T_w$ and a stride $T_h$. The data subset for pre-training uses $T_w=30s$, $T_h=30s$, the subset for fine-tuning and the test set use $T_w=30s$, $T_h=15s$. Each segment's annotation is determined using the Illgraben event catalogue, which is split into training and test set with a ratio of approx. 70/30. Linear detrend is applied to the seismic signal before it is transformed into a log-compressed spectrogram with window length of 2.56\ s and stride of 0.08\ s. No data augmentation is applied. As encoder we use a single-channel variant of ResNet18 \citep{heDeepResidualLearning2015}, without the final linear layer. The output of the encoder is a 512-dimensional vector, which is then passed through the encoder head, consisting of a MLP with a hidden layer of size 512, batch normalization and ReLU non-linearity. The encoder head's output size is $d=128$ and L2 normalized. During fine-tuning, the same encoder head architecture (with random initialization) is used as a classification head with an output size equal to the number of classes $N_c$.

In the supervised training we train encoder and classification head jointly using cross-entropy loss. 

For semi-supervised training (\textit{all+SC} and \textit{one-station+SC}), we train the encoder and encoder head with contrastive loss, then for fine-tuning we replace the encoder head with a classification head and train the classification head with cross-entropy loss while keeping the encoder weights fixed. In every epoch we first train with contrastive loss, then fine-tune the classification head. For optimization we use SGD with momentum 0.9 and weight decay $10^{-4}$ and a batch size of 128. The temperature coefficient $\tau$ is set to 0.1. We use a cosine annealing scheduler. In our experiments the edge weights of the information graph are 1. 

To counter class-imbalance, each batch contains the same number of examples for each class. During semi-supervised training the non-annotated data out-weights the annotated data by a factor of approx. 4.5 in each batch. We select our model based on a validation set which is 20\% of the training set, except for the one-station+SC experiment. Here, we select model from the last epoch, since model selection on the one-station subset would deteriorate the generalization effect. 
More details, e.g. the code and hyperparameters for individual experiments will be made available on the paper's project page.

\subsection{Details to Random Forest Classifier}
Following\citet{Provost2017} and \citet{Wenner2020} we computed a total of 55 signal characteristics in the time and frequency domain, e.g., information on the signal form and dominant frequencies. For a complete description of the chosen features see \citet{Wenner2020}. We performed a three-fold-cross-validation grid search to optimize classifier performance. Final parameters are presented in Table \ref{tab:rf}.

\begin{table}[t]
\caption{Random forest parameters}
\label{tab:rf}
\begin{center}
\begin{tabular}{ll}
Number of trees         &400 \\
Split quality measure   &Gini criterion \\
Minimum number of samples required to be a leaf node  & 1 \\
Minimum number of samples for an internal node to be split & 2 \\
\end{tabular}
\end{center}
\end{table}

\end{document}

%% file: math_commands.tex
\usepackage{amsmath,amsfonts,bm}

\def\eqref#1{equation~\ref{#1}}

\def\1{\bm{1}}

\def\va{{\bm{a}}}

\def\vz{{\bm{z}}}

\def\mA{{\bm{A}}}
\def\mB{{\bm{B}}}

\DeclareMathAlphabet{\mathsfit}{\encodingdefault}{\sfdefault}{m}{sl}
\SetMathAlphabet{\mathsfit}{bold}{\encodingdefault}{\sfdefault}{bx}{n}

\def\emB{{B}}

\newcommand{\R}{\mathbb{R}}

